\documentclass[acmtog, dvipsnames,screen,nonacm]{acmart}  

\usepackage{booktabs} 
\usepackage{multirow}
\usepackage{makecell}
\usepackage{listings}
\usepackage{xcolor}
\usepackage{svg}
\usepackage[percent]{overpic}
\usepackage{tikz}
\usepackage{xcolor}
\usepackage{hyperref}

\definecolor{jointYellow}{HTML}{FEE695}

\usepackage{tabularx,booktabs} 
\usepackage{hyphenat,balance}
\usepackage[fleqn,tbtags]{mathtools}
\usepackage{overpic}  
\usepackage{wrapfig}
\usepackage{contour}
\usepackage{microtype}
\usepackage{multirow}
\usepackage{colortbl}
\usepackage{amsmath}
\usepackage{bbm}
\usepackage{textcomp}
\usepackage{enumitem}
\setitemize{noitemsep,topsep=0pt,parsep=0pt,partopsep=0pt}
\usepackage{derivative}
\usepackage{float}

\usepackage{graphicx}
\graphicspath{{./images/}}

\definecolor{mypink1}{RGB}{205,115,114}

\usepackage[linesnumbered,ruled,lined]{algorithm2e} 
\SetAlFnt{\small}
\SetAlCapFnt{\small}
\SetAlCapNameFnt{\small}
\SetAlCapHSkip{0pt}

\usepackage{epigraph}
\usepackage{algpseudocode}
\usepackage{mathrsfs}
\usepackage{makecell}

\usepackage{amsmath,amsfonts,bm}

\def\1{\bm{1}}

\DeclareMathAlphabet{\mathsfit}{\encodingdefault}{\sfdefault}{m}{sl}
\SetMathAlphabet{\mathsfit}{bold}{\encodingdefault}{\sfdefault}{bx}{n}

\newcommand{\name}{\emph{AnyAct}\xspace}

\acmSubmissionID{1181}

\begin{document}

\title{\name: Towards Human Reenactment of Character Motion From Video}

\author{Liuhan Chen}
\authornote{Both authors contributed equally to this research.}
\affiliation{%
  \institution{Peking University}
  \city{Peking}
  \country{China}
}
\email{liuhanchen@stu.pku.edu.cn}

\author{Lei Zhong}
\authornotemark[1]
\affiliation{%
  \institution{Nankai University}
  \city{Tianjin}
  \country{China}
}
\email{leizhong.nankai@gmail.com}

\author{Jiawei Wang}
\affiliation{%
\institution{ShanghaiTech University}
\city{Hongkong}
\country{China}
}
\email{jiaweiwang0222@gmail.com}

\author{Qing Shuai}
\affiliation{%
\institution{Zhejiang University}
\city{hangzhou}
\country{China}
}
\email{s_q@zju.edu.cn}

\author{Li Yuan}
\affiliation{%
\institution{Peking University}
\city{Peking}
\country{China}
}
\email{yuanli-ece@pku.edu.cn}

\author{Leidong Fan}
\affiliation{%
\institution{Pengcheng Laboratory}
\city{Shenzhen}
\country{China}
}
\email{leidongfan@gmail.com}

\author{Qing Li}

\authornote{Corresponding author.}
\affiliation{%
\institution{Pengcheng Laboratory}
\city{Shenzhen}
\country{China}
}
\email{lqing900205@gmail.com}

\author{Kanglin Liu}
\affiliation{%
\institution{Pengcheng Laboratory}
\city{Shenzhen}
\country{China}
}
\email{max.liu.426@gmail.com}


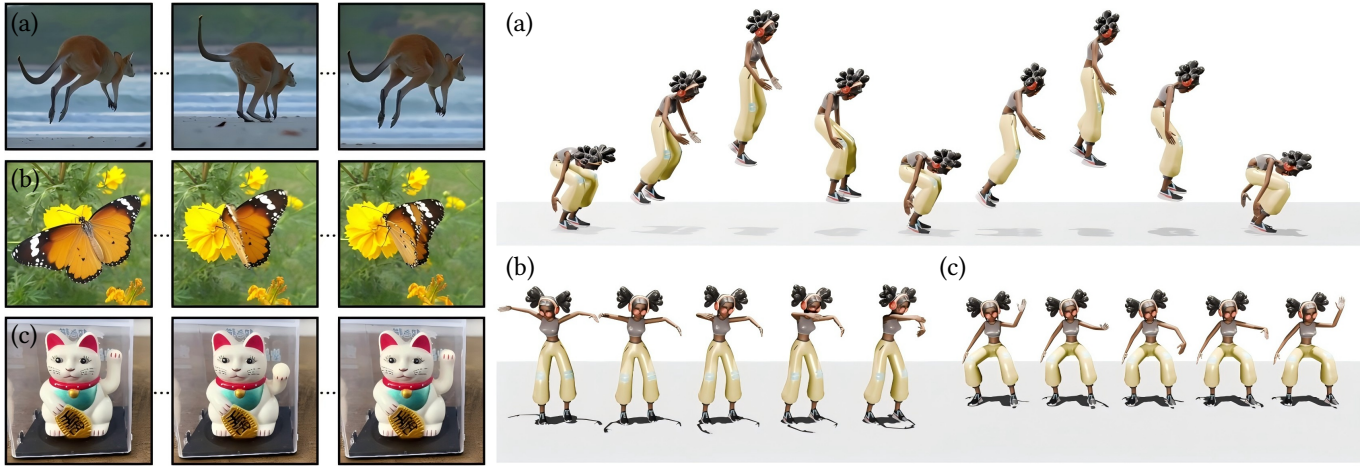
\begin{teaserfigure}
    \centering
    \begin{overpic}[width=\textwidth]{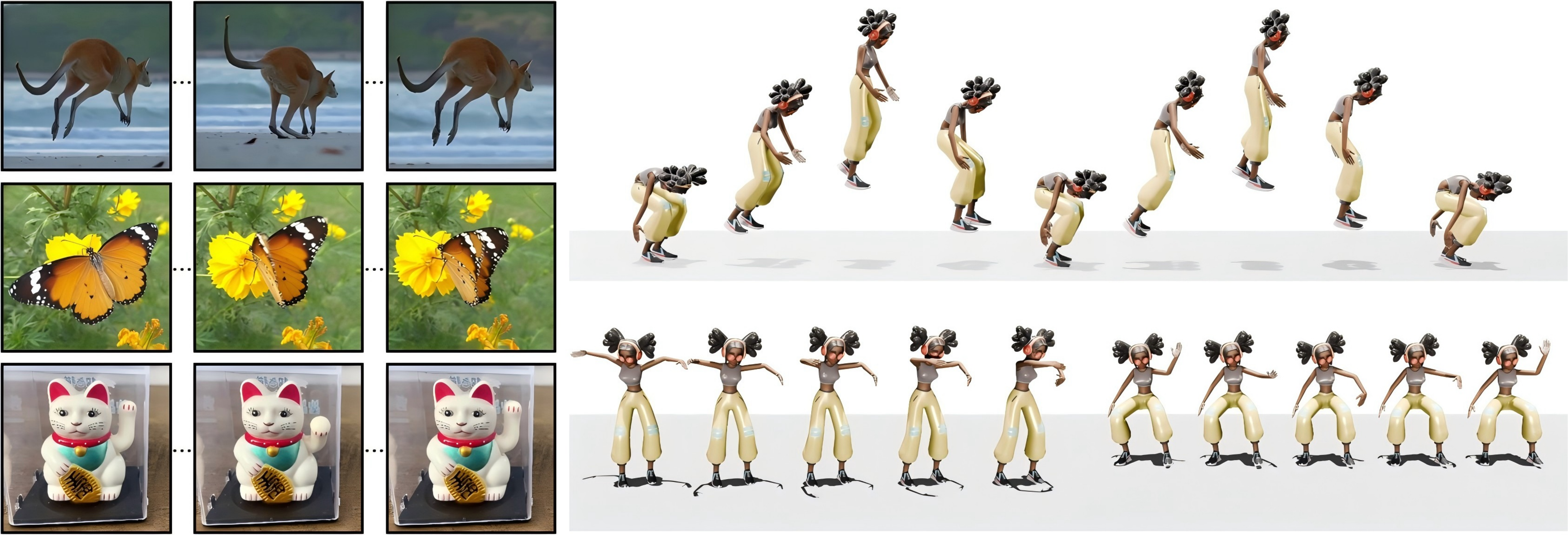}
        \put(0.5,32){\large (a)}
        \put(37,32){\large (a)}
        \put(0.5,20.5){\large (b)}
        \put(37,14){\large (b)}
        \put(0.5,9){\large (c)}
        \put(69,14){\large (c)}
    \end{overpic}
    \caption{\textbf{Human reenactment from reference videos.}
    Given monocular videos of non-human characters with diverse topologies, \name\ reinterprets their characteristic motion patterns as plausible human performances rather than reproducing their source structures literally. Shown here are reenactments of (a) kangaroo-like jumping, (b) butterfly-like wing flapping, and (c) the periodic paw motion of a beckoning cat.
    }
    \label{fig:teaser}
\end{teaserfigure}

\begin{abstract}
We study the problem of directly deriving an initial human reenactment from a monocular video of a non-human character. Our goal is not to reconstruct the source character itself but to reinterpret its motion as a plausible and editable human performance for downstream animation authoring. This task is challenging because existing video-based motion capture methods are largely restricted to human-centric structural spaces, while motion retargeting methods typically require structured 3D source motions and known source topologies.
Our key insight is that sparse local articulated motion cues can preserve essential dynamics across large structural differences, providing a stable bridge from character video to human reenactment. Based on this observation, we propose \name, which formulates character-video-driven human reenactment as conditional human motion generation from transferable sparse local 2D articulated motion. To make this practical, we introduce three key designs: human-motion-only supervision via augmented 3D-to-2D projection, progressive 3D-to-2D training to alleviate conditioning ambiguity, and global-local motion decoupling for reliable local motion control.
We further construct a benchmark primarily covering diverse non-human character videos. Experiments on the benchmark show that \name produces high-fidelity initial human reenactments that preserve the essential dynamics of the characters in reference videos, and further ablation studies validate the effectiveness of its core designs.
\emph{Results are best visualized through} \url{https://anyact.github.io}.
\end{abstract}

%
%


\maketitle

\section{Introduction}
Animators often draw inspiration from non-human motion references when crafting motion for human or human-like characters. For example, to create a shot with kangaroo-like jumping dynamics after observing a kangaroo-jumping video, an artist should not literally reproduce the structure of the animal but reinterpret its characteristic motion patterns into a human performance. In practice, this process typically starts with an initial human reenactment that captures character-specific dynamics from the reference video while maintaining the plausibility and editability of the human motion. However, current authoring pipelines provide little support for deriving such an initial reenactment from non-human motion references, making this step largely manual, time-consuming, and heavily dependent on artistic expertise.

This naturally motivates a critical question: \emph{Can we design a framework to streamline the derivation of a plausible initial human reenactment directly from a monocular video recording the specific motion of a non-human character?} Our goal is not to faithfully reconstruct the source character in its original form but to reinterpret its motion in the human domain as an editable starting point for downstream authoring. The challenge for such a framework lies in handling diverse and non-unified source structures: the body structures of source characters often differ significantly from those of humans, and non-human characters themselves exhibit wide variability in structural forms and motion patterns. Besides, monocular videos captured in unconstrained environments limit access to accurate 3D geometric information of non-human characters.

Current methods either focus only on the human domain or assume prior access to sufficient 3D geometric information of non-human characters. Specifically, Video-based motion capture approaches~\cite{TOG2020_motionet, SA2024_world, CVPR2025_d3human, ICCV2025_GENMO} reconstruct 3D human motion from monocular videos, but they are largely designed for human-centric inputs and fixed human structural spaces~\cite{TOG2015_smpl, CVPR2019_SMPLXexpressive}. As a result, they are not well suited for non-human character videos, where the observed motion may be expressed through body structures that are very different from those of humans. Motion retargeting methods~\cite{SA2024_walkthedog, arxiv2026_palum, SA2023_same}, in contrast, can transfer motion across different skeletal structures, but they typically rely on structured 3D source motions with known source 3D topologies. Such assumptions are difficult to satisfy for a single in-the-wild video of a non-human character. Consequently, deriving an initial human reenactment directly from a non-human character video remains largely unsupported by existing formulations.

To address this gap, the key is not to faithfully recover the source character or its full-body structure, but to focus on the dynamic impression conveyed by local articulated motion, which remains more consistent across large structural differences. As illustrated in Fig.~\ref{fig:motivation}, although the motions of a non-human character (e.g., a jumping kangaroo) and its human reenactment may differ substantially in morphology and topology, their sparse local articulated movements still carry essential and similar dynamic tendencies. This suggests that local sparse motion patterns provide a more stable bridge between monocular character video observations and human reenactment than source-specific full-body structure.

Building on this observation, we propose \name, a framework that formulates character-video-driven reenactment as conditional human motion generation from transferable local sparse motion cues. Concretely, \name extracts sparse 2D motion patterns from monocular character videos to condition a human motion generator for initial reenactment synthesis. Since paired non-human character videos and 3D human reenactment motions are unavailable, we train the framework using only human motion data with projected 2D conditions. Because sparse 2D cues provide weak and ambiguous supervision for 3D motion generation, we train the model progressively from reliable 3D motion supervision to projected 2D conditions that better match the target setting. Finally, because reference videos more directly constrain local dynamics than human-compatible global displacement, we decouple the global trajectory from the local motion. Together, these designs enable \name to directly derive strong initial human reenactments from monocular non-human character videos without requiring source 3D topology or complex skeletal correspondence.

To systematically evaluate our \name, we construct a benchmark primarily containing diverse non-human character videos from real-world sources, covering animals and other non-human entities with varied motion patterns and structural forms. Across this benchmark, \name consistently produces high-fidelity initial human reenactments that preserve the essential dynamics of the characters in videos. Extensive experiments and ablation studies validate the effectiveness of our framework and its core designs. In summary, the principal contributions of this work include:
\begin{itemize}
    \item We introduce a new task, \textit{human reenactment from character motion video}, and construct a benchmark with evaluation protocols for this setting.

    \item We propose \name, a framework that formulates human reenactment from character video as conditional human motion generation from transferable local sparse 2D motion cues, without requiring source 3D topology or complex skeletal correspondence.

    \item We introduce three key designs that make this formulation practical: human-motion-only supervision via augmented 3D-to-2D projection, progressive 3D-to-2D training to reduce motion ambiguity, and global-local motion decoupling for reliable local motion control.
\end{itemize}

\begin{figure}[t]
    \centering

    \begin{tikzpicture}
        \node[anchor=south west, inner sep=0] (img) at (0,0)
        {\includegraphics[width=\linewidth]{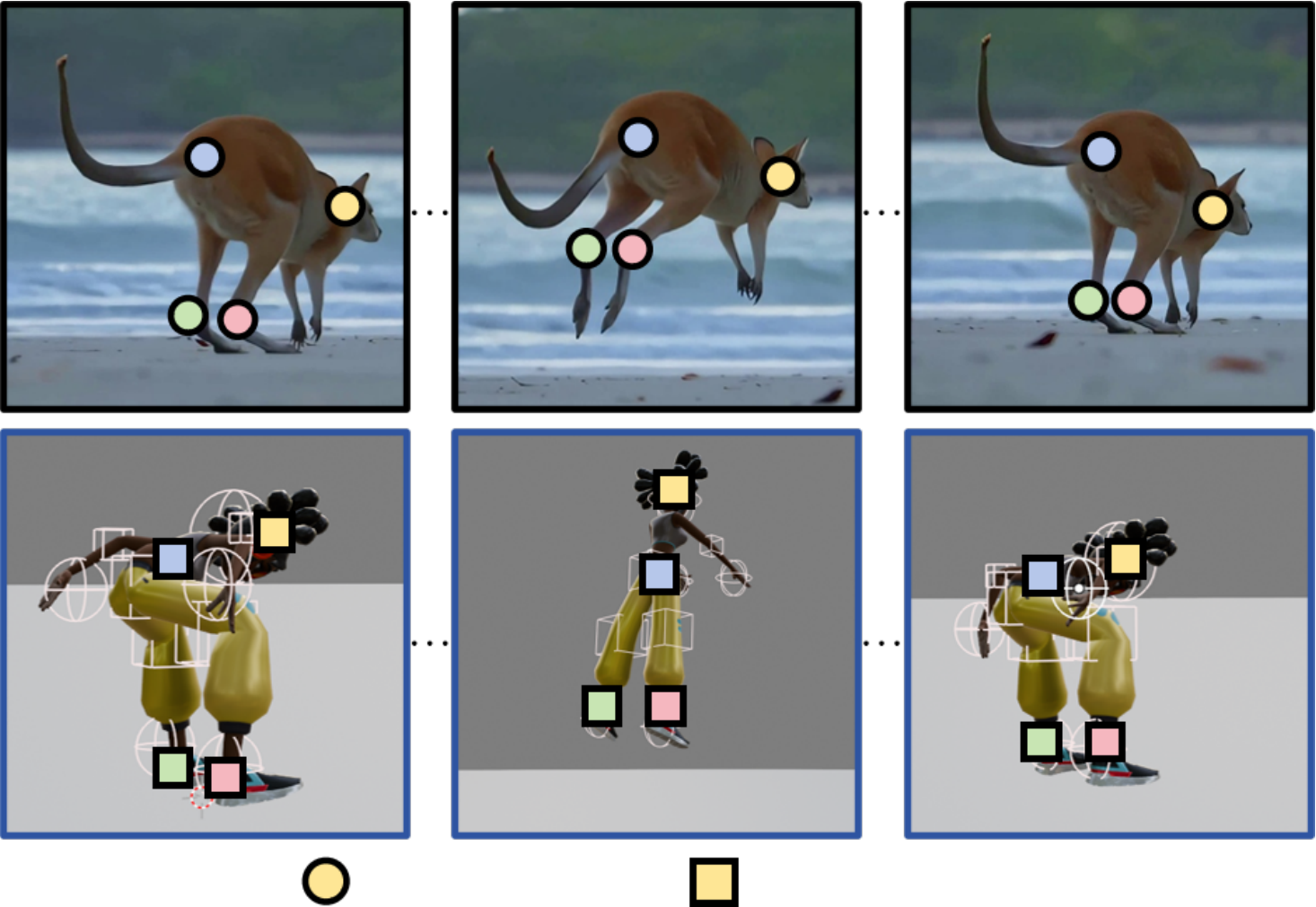}};

        \begin{scope}[x={(img.south east)}, y={(img.north west)}]

            \node[anchor=west, font=\fontsize{7.5}{8}\selectfont\bfseries]
                at (0.0,0.026) {Kangaroo Joint};

            \node[anchor=west, font=\fontsize{7.5}{8}\selectfont\bfseries]
                at (0.322,0.026) {Human Joint};

        \end{scope}
    \end{tikzpicture}
    \vspace{-6mm}
     \caption{\textbf{Sparse local motion cues across structural differences.} Top: sparse local joint movements extracted from a jumping kangaroo video. Bottom: the corresponding sparse local joint movements of its human reinterpretation, obtained via time-consuming manual design in Blender. Although they differ substantially in full-body structure, their local sparse articulated motion still conveys similar dynamic tendencies.}
     \label{fig:motivation}
\end{figure}
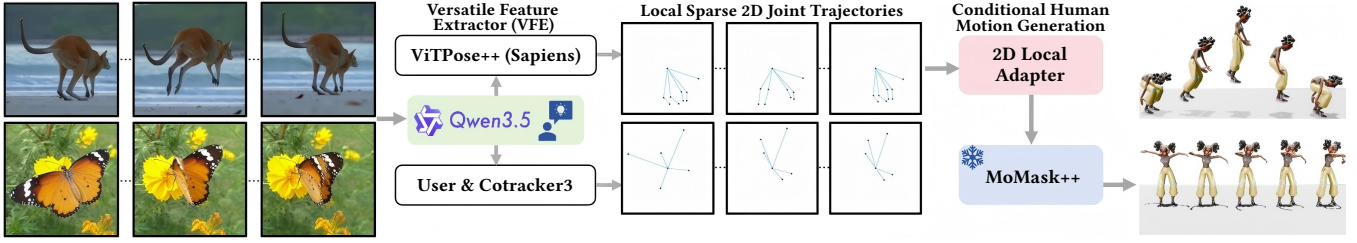
\begin{figure*}[!tb]
    \centering
    \begin{tikzpicture}
        \node[anchor=south west, inner sep=0] (img) at (0,0)
        {\includegraphics[width=0.99\linewidth]{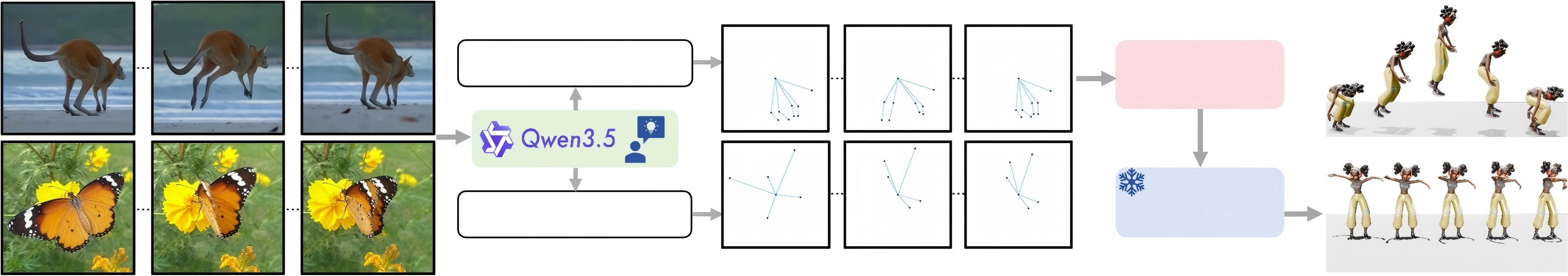}};
        \begin{scope}[x={(img.south east)}, y={(img.north west)}]

            \node[anchor=west, font=\fontsize{6.5}{8}\selectfont\bfseries]
                at (0.309,0.965) {Versatile Feature};

            \node[anchor=west, font=\fontsize{6.5}{8}\selectfont\bfseries]
                at (0.314,0.895) {Extractor (VFE)};

            \node[anchor=west, font=\fontsize{6.5}{8}\selectfont\bfseries]
                at (0.468,0.955) {Local Sparse 2D Joint Trajectories};

            \node[anchor=west, font=\fontsize{6.5}{8}\selectfont\bfseries]
                at (0.700,0.965) {Conditional Human};

            \node[anchor=west, font=\fontsize{6.5}{8}\selectfont\bfseries]
                at (0.703,0.895) {Motion Generation};

            \node[anchor=west, font=\fontsize{7}{7.5}\selectfont\bfseries]
                at (0.297,0.766) {ViTPose++ (Sapiens)};

            \node[anchor=west, font=\fontsize{7}{7.5}\selectfont\bfseries]
                at (0.302,0.225) {User \& Cotracker3};

            \node[anchor=west, font=\fontsize{7.2}{8}\selectfont\bfseries]
                at (0.73,0.78) {2D Local};

            \node[anchor=west, font=\fontsize{7.2}{8}\selectfont\bfseries]
                at (0.732,0.68) {Adapter};

            \node[anchor=west, font=\fontsize{7.2}{8}\selectfont\bfseries]
                at (0.725,0.26) {MoMask++};

        \end{scope}
    \end{tikzpicture}

    \vspace{-1.5mm}
    \caption{Given reference videos of characters, \name first extracts local sparse 2D joint trajectories as transferable motion cues from the input video using our model-ensemble-based Versatile Feature Extractor (VFE).
    These cues are then injected into a human motion generator (MoMask++) through the ControlNet-like 2D Local Adapter (2D-LA) to produce the initial human reenactments that follow the observed character dynamics.}
    \label{fig:pipeline}
\end{figure*}

\section{Related Work}
\paragraph{Video-based 3D Motion Capture.}
Video-based 3D motion capture aims to recover pre-defined 3D structural sequences of moving subjects from videos.
For monocular human motion reconstruction~\cite{CVPR2021_pymaf, CVPR2021_agora}, early methods typically first estimate a fixed-topology 2D human skeleton in image space~\cite{PAMI2019_openpose} and then recover 3D parametric body representations, mainly SMPL/SMPLX~\cite{TOG2015_smpl}, through optimization~\cite{ECCV2016_keepitsmpl, CVPR2019_SMPLXexpressive}.
More recent approaches~\cite{CVPR2025_prompthmr, CVPR2025_d3human} directly predict 3D skeletons~\cite{ICCV2023_humans, CVPR2026_FMPose3D} or parameters~\cite{NIPS2023_smpler, SA2024_world} in an end-to-end manner~\cite{CVPR2024_chatpose, CVPR2024_tokenhmr}, benefiting from stronger feature backbones~\cite{ICCV2023_motionbert, arxiv2025_sambody, arxiv2026_DuoMo, arxiv2026_Onlinehmr}. Several systems further improve reconstruction quality in controlled indoor settings by exploiting multi-view human videos~\cite{arxiv2021_easymocap, SG2022_multinb}.

Beyond humans, several works extend video-based 3D motion capture to animals~\cite{3DV2024_farm3d} and other articulated objects~\cite{CVPR2024_4dfly, arxiv2026_necromancer}.
In particular, recent studies recover articulated 3D animal motion from in-the-wild videos~\cite{CVPR2023_magicpony, ECCV2024_ponymation, NIPS2025_web} by combining species-specific keypoint detection~\cite{NIPS2022vitpose, PAMI2023_vitpose++} with parametric model fitting~\cite{CVPR2017_3dmenagerie}.
Recently, DancingBox~\cite{CHI2026_DancingBox} moves toward video-driven transfer from the dynamics of non-human physical proxies to human motion, but it relies on constrained capture setups with  explicitly separable ground markings, user interactions, and uncertain per-sample hyperparameter adjustments, which limit its applicability to in-the-wild character videos.
Another concurrent work (still without released model weights), MoCapAnything-V2~\cite{arxiv2025_mocapanything,arxiv2026_mocapanythingv2}, utilizes deterministic modeling to directly retarget animal motions in videos to the human skeleton. However, compared to generative methods, it ignores the motion plausibility and struggles to incorporate external controls.

Overall, prior video-based motion capture methods focus on recovering motion within pre-defined structural domains, whereas our goal is to reinterpret the dynamics of versatile characters with different topologies in videos as human motion.

\paragraph{Motion Retargeting Across Topologies.}
Motion retargeting aims to transfer motion across characters with different skeletal structures or morphologies.
Early methods typically rely on manually designed kinematic constraints~\cite{SG1998_Retargetting} and space-time optimization~\cite{SG1999_hierarchical, TOG2005_physically}, while later learning-based approaches enable more flexible and data-driven motion transfer~\cite{SA2018_variational, CVPR2018_neural}.
More recent work extends retargeting beyond shared skeleton structures by learning correspondences or common latent spaces across different topologies.
For example, graph-based methods embed skeletons into shared representations for cross-skeleton motion transfer~\cite{ToG2020_skeleton, SA2023_same}, while body-part-based retargeting further improves flexibility by treating body parts as transferable units rather than requiring strict whole-body correspondence~\cite{hu2023pose}.
Subsequent approaches address larger morphology gaps, such as human-to-quadruped transfer~\cite{SA2024_walkthedog, SG2024_dog}, through shared codebooks, phase manifolds, context matching, or sparse correspondence modeling~\cite{SA2023_mocha, SA2025_motion2motion}.

Recent advances further improve robustness and generalization by incorporating contact-aware constraints or adopting unsupervised and generative formulations~\cite{arxiv2025_reconform, TOG2025_kinematic, arxiv2025_moreflow, arxiv2026_palum}.
Closely related to our setting, video-based retargeting methods such as TransMoMo~\cite{yang2020transmomo} and MoCaNet~\cite{zhu2022mocanet} begin to transfer motion directly from monocular videos without explicit 3D reconstruction.
However, existing retargeting methods still assume structured source 3D motions, human/body-centric video domains, or explicit source skeleton definitions before transfer, whereas our setting starts directly from categorically versatile monocular character motion videos without requiring source 3D topologies.

\paragraph{Controllable Human Motion Generation}
Data-driven generative models learn realistic human motion priors~\cite{3DV2019Language2pose, ToG2020_unpaired, TOG2022_deepphase}, and many recent works further improve controllability through conditional generation, most notably with text guidance~\cite{ICLR2023_MDM, CVPR2023_LDM, CVPR2023_generating, CVPR2024_momask, arxiv2026_kimodo}.
More recent methods further incorporate additional control signals such as style~\cite{TOG2022_motion, CVPR2024_arbitrary, ICCV2025_stylemotif} and trajectory~\cite{SA2024_flexible, ICLR2024_omnicontrol, ICCV2025_maskcontrol,TOG2025_sketch2Anim}, either in a zero-shot manner~\cite{SA2024_monkey} or through adapter-based designs~\cite{ECCV2024_smoodi}.

Related to topology-aware character animation, recent work also explores motion generation for arbitrary skeletons~\cite{SG2025_Anytop}, but such methods require skeletal structure as input rather than conditioning on character video.
Although some recent works begin to process video and motion in a shared space~\cite{arxiv2026_comovi, ICLR2026_echomotion} based on video generation models~\cite{arxiv2025_wan}, they remain limited to videos containing human motion.
In contrast, our work bridges the gap between disparate character observations and human motion synthesis by identifying local sparse 2D joint trajectories~\cite{ICCV2025_motion23} as a unified motion cue.
This allows our framework to effectively transfer dynamics from categorically versatile character videos to 3D human motions while being trained exclusively on standard humanoid motion capture data.

\section{Method}
\label{sec:method}

\subsection{Overview}
\label{Method_Subsec: Overview}
Given an $N$-frame monocular video $\boldsymbol{V}=\{\boldsymbol{v}_n\}_{n=1}^N$ of a character, our goal is to generate a 3D human motion sequence $\boldsymbol{M}=\{\boldsymbol{m}_n\}_{n=1}^N$ that reenacts the observed dynamics. For inference, as illustrated in Fig.~\ref{fig:pipeline}, \name first extracts local sparse 2D joint trajectories $\boldsymbol{J}=\{\boldsymbol{j}_n\}_{n=1}^N$ as transferable motion cues from the input video using our Versatile Feature Extractor (VFE). Then, following ControlNet, a trainable copy branch (2D-LA) processes these cues and injects them into the frozen base motion generator, MoMask++~\cite{NIPS2025_snapmogen} by layer-wise residuals addition, yielding an initial human reenactment that follows the observed character dynamics.

This formulation involves two main challenges: extracting comparable sparse motion cues from character videos with diverse topologies and training a reliable motion control branch to encode and inject such cues. The first is addressed by our designed VFE (Sec.~\ref{Method_Subsec: Feature Representation and Extraction}), while the second is handled by our proposed motion condition learning strategy (Sec.~\ref{Method_Subsec: Progressive Motion Condition Control Training}), which tackles the lack of paired supervision, 2D-to-3D condition ambiguity, and unreliable global root motion.

\subsection{Versatile Motion Feature Extraction}
\label{Method_Subsec: Feature Representation and Extraction}

Our goal is to extract motion cues that remain comparable across diverse character topologies and are usable for downstream human motion generation. To this end, we define a unified sparse 2D motion representation shared by most input characters.

Specifically, each character is decomposed into five local components: four limbs and one torso-head segment, with each component represented by at most two joints. These joints correspond to human joints with key motion semantics, including the root and head, left elbow and wrist, right elbow and wrist, left knee and ankle, and right knee and ankle. Missing joints are handled through zero-padding and valid masking. This representation remains general across diverse character morphologies while preserving the essential local motion cues for human reenactment. 

To obtain this representation from diverse videos, we combine category-specific estimators with a flexible fallback strategy. We first use ViTPose++~\cite{PAMI2023_vitpose++} to extract keypoint trajectories in the AP-10K format~\cite{NIPS2021_ap10k} from a wide range of animal characters.
To generalize our framework to humanoid characters, we also incorporate an advanced human-domain estimator, Sapiens~\cite{ECCV2024_sapiens} to extract keypoint trajectories in the COCO-17 format. For out-of-domain characters, we instead allow lightweight user annotation of semantically meaningful keypoints and track them temporally with CoTracker3~\cite{ICCV2025_cotracker3}. To improve automation, we further use Qwen3.5-9B~\cite{arxiv2026_qwen3.5} to route each input video to the appropriate extraction path.
Finally, to reduce scale and translation ambiguity in monocular videos, we anchor the root at $\boldsymbol{0}$ and apply global $L_2$-normalization to the remaining joints. The output of VFE is a standardized sequence of local sparse 2D joint trajectories, which serves as the transferable motion condition for subsequent human motion generation.
More details about the feature extraction are provided in the Appendix.

\subsection{Motion Condition Learning}

\begin{figure}[t] 
    \centering

    \begin{tikzpicture}
        \node[anchor=south west, inner sep=0] (img) at (0,0)
        {\includegraphics[width=\linewidth]{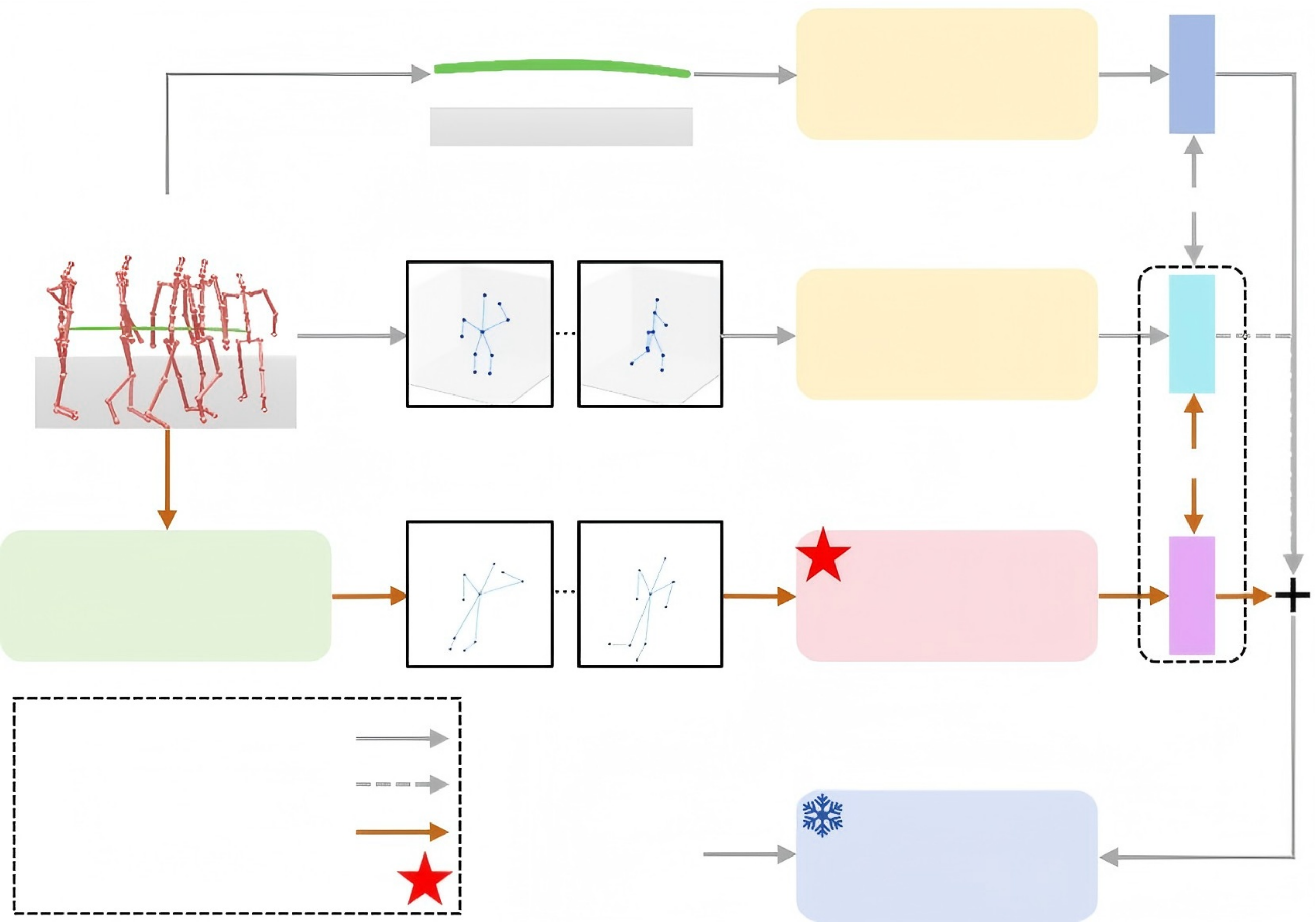}};

        \begin{scope}[
            x={(img.south east)},
            y={(img.north west)},
            font=\footnotesize
        ]

        \node[anchor=center, font=\footnotesize\bfseries] at (0.13,0.76) {3D Human Motion};

        \node[anchor=center, font=\footnotesize\bfseries] at (0.130,0.375) {Augmented 3D-};
        \node[anchor=center, font=\footnotesize\bfseries] at (0.130,0.325) {to-2D Projection};

        \node[anchor=center, font=\footnotesize\bfseries] at (0.428,0.999) {Global 3D};
        \node[anchor=center, font=\footnotesize\bfseries] at (0.428,0.959) {Root Trajectory};

        \node[anchor=center, font=\footnotesize\bfseries] at (0.428,0.782) {Local Sparse 3D};
        \node[anchor=center, font=\footnotesize\bfseries] at (0.428,0.742) {Joint Trajectories};

        \node[anchor=center, font=\footnotesize\bfseries] at (0.428,0.500) {Local Sparse 2D};
        \node[anchor=center, font=\footnotesize\bfseries] at (0.428,0.460) {Joint Trajectories};

        \node[anchor=center, font=\footnotesize\bfseries] at (0.715,0.938) {3D Global};
        \node[anchor=center, font=\footnotesize\bfseries] at (0.715,0.888) {Adapter};

        \node[anchor=center, font=\footnotesize\bfseries] at (0.715,0.658) {3D Local};
        \node[anchor=center, font=\footnotesize\bfseries] at (0.715,0.608) {Adapter};

        \node[anchor=center, font=\footnotesize\bfseries] at (0.715,0.375) {2D Local};
        \node[anchor=center, font=\footnotesize\bfseries] at (0.715,0.325) {Adapter};

        \node[anchor=center, font=\footnotesize\bfseries] at (0.725,0.075) {MoMask++};

        \node[anchor=center, font=\footnotesize\bfseries] at (0.905,0.790) {$L_{O}$};
        \node[anchor=center, font=\footnotesize\bfseries] at (0.905,0.505) {$L_{3D}$};

        \node[anchor=center, font=\footnotesize\bfseries] at (0.133,0.206) {3D \& 2D Training};
        \node[anchor=center, font=\footnotesize\bfseries] at (0.133,0.153) {3D Training Only};
        \node[anchor=center, font=\footnotesize\bfseries] at (0.133,0.100) {2D Training Only};
        \node[anchor=center, font=\footnotesize\bfseries] at (0.157,0.047) {Adapter for Inference};

        \node[anchor=center, font=\footnotesize\bfseries] at (0.457,0.075) {$L_{MoMask++}$};

        \end{scope}
    \end{tikzpicture}
    \vspace{-5.5mm}
    \caption{\textbf{Motion Condition Learning.}
    We learn reliable motion control for our \name using only human motion data.
    This is achieved by our proposed augmented 3D-to-2D projection for providing paired supervision, 
    progressive 3D-to-2D training to alleviate conditioning ambiguity, and global-local motion decoupling 
    for suppressing unreliable global root motion.}
    
    \label{fig:Training_method}
\end{figure}
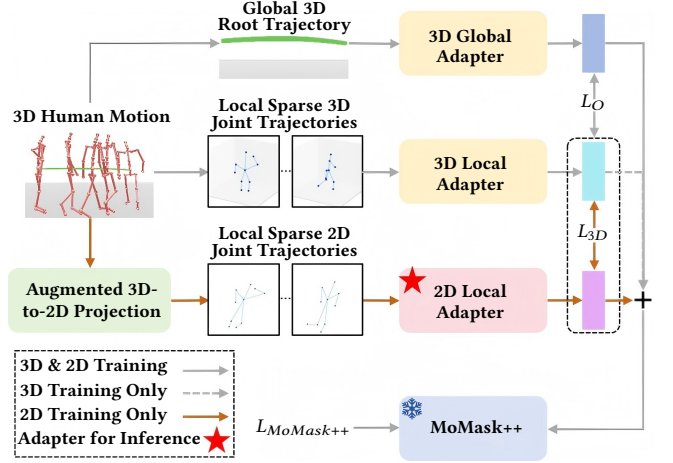

\label{Method_Subsec: Progressive Motion Condition Control Training}

As shown in Fig.~\ref{fig:Training_method}, we learn reliable motion control using only human motion data. This is achieved by our proposed augmented 3D-to-2D projection for providing paired supervision, progressive 3D-to-2D training to alleviate conditioning ambiguity, and global-local motion decoupling to suppress unreliable global root motion.

\paragraph{Augmented 3D-to-2D Projection.}

Since paired non-human video cues and 3D human reenactment motions are unavailable, we construct condition-motion pairs entirely from 3D human motion data. Specifically, we project 3D human motions onto the 2D plane and convert them into the same sparse local representation used at inference time, including joint selection, masking, and normalization. To better approximate monocular character videos captured under diverse and uncertain camera settings, we sample virtual cameras with random viewpoints, distances, and trajectories.

However, projected human motion still differs from real character video cues in several aspects. Characters may exhibit different body proportions and inherent postures, while joint estimation and point tracking can introduce noise and jitter. To narrow this gap, we augment the projected 2D joint sequences with random symmetric limb scaling, global rotation, and Gaussian perturbation. These augmentations improve the robustness of our motion condition learning to diverse morphologies, postural variations, and extraction errors. More details about the feature extraction are listed in the Appendix.

\paragraph{Progressive 3D-to-2D Training.}
Since multiple plausible 3D motions may correspond to the same 2D observation, conditioning on sparse 2D motion cues is inherently under-constrained. Directly predicting reliable 3D joint cues for versatile characters from monocular videos is also difficult in practice. Therefore, we adopt a progressive 3D-to-2D training scheme to alleviate the ambiguity.

Specifically, we first train a 3D Local Adapter (\textit{3D-LA}) using normalized 3D joint sequences as conditions. We then train a 2D Local Adapter (\textit{2D-LA}) and align its output with that of the stronger 3D-conditioned branch through an $L_2$ alignment loss:
\begin{equation}
\mathcal{L}_{3D} = \left\| \textit{3D-LA}(\boldsymbol{J}_{3D}) - \textit{2D-LA}(\boldsymbol{J}_{2D}) \right\|_2,
\end{equation}
where $\boldsymbol{J}_{3D}$ denotes the 3D joint sequence and $\boldsymbol{J}_{2D}$ its 2D projection. In this way, the 2D branch inherits structural knowledge from the stronger 3D-conditioned branch, making 2D control more 3D-aware while still requiring only sparse 2D cues at inference time.

\paragraph{Global-Local Decoupled Learning.}
Our sparse motion conditions mainly encode local motion cues, removing the global root movements that are difficult to accurately obtain form monocular character videos. If a single local branch is forced to explain both local dynamics and global trajectory, local motion conditioning becomes entangled with ill-posed global motion during inference. This degrades reenactment fidelity and reduces the flexibility of subsequent trajectory control. We therefore explicitly decouple local motion conditioning from global trajectory modeling during training.

Specifically, given a 3D human motion, the LA branch is responsible for processing local motion cues, while we introduce an auxiliary 3D Global Adapter (\textit{3D-GA}) to model the global root trajectory $\boldsymbol{T}$. To encourage the two branches to encode complementary rather than redundant information, we further impose an regularizer:
\begin{equation}
\mathcal{L}_{O} = \mathcal{P}\big(\textit{3D-GA}(\boldsymbol{T}), \textit{LA}(\boldsymbol{J})\big),
\end{equation}
where $\mathcal{P}(\cdot,\cdot)$ denotes the squared cosine similarity between the global and local condition features. This orthogonal regularization reduces entanglement between the two branches and encourages LA to focus on local motion control information.

The $\mathcal{L}_{O}$ is incorporated into both training stages. During 3D training, we optimize \textit{3D-LA} and \textit{3D-GA} jointly, expressed as:
\begin{equation}
\mathcal{L}_{\text{Total-3D}}=\mathcal{L}_{\text{MoMask++}} + \lambda_1 \mathcal{L}_{O},
\end{equation}
During 2D training, we freeze the \textit{3D-LA} and optimize \textit{2D-LA} and \textit{3D-GA} jointly, which can be expressed as:
\begin{equation}
\label{total_loss}
\mathcal{L}_{\text{Total-2D}}=\mathcal{L}_{\text{MoMask++}} + \lambda_1 \mathcal{L}_{O} + \lambda_2 \mathcal{L}_{3D},
\end{equation}
where $\mathcal{L}_{\text{MoMask++}}$ is the original training loss of the base generator (detailed in Appendix). 
At inference time, we retain only the 2D-LA branch and discard the auxiliary training-time modules.

\section{Results and Discussion}
\label{sec:results}

\subsection{Experimental Setting}
\paragraph{Dataset}
For training, we adopt the SnapMoGen~\cite{NIPS2025_snapmogen} dataset, which encompasses 43.7 hours of high-quality motion data with detailed textual descriptions.
This dataset serves as the official training set for our Text-to-Motion base model.
For evaluation, we collected 160 real-world motion videos. These primarily feature non-human characters while retaining a small number of humanoid character videos to assess generalization, each paired with a manually crafted concise prompt for human reenactment (e.g., "He jumps forward" or "The person walks forward").
More details about the evaluation data are provided in the Appendix.

\paragraph{Implementation Details}
All adapters in our work are ControlNet-like networks, featuring a full copy of the base model and zero-initialized output layers.
We train them for a total of 30 epochs with a learning rate of 2e-4 and a batch size of 64.
For better motion condition learning, we utilize an advanced LLM, Qwen3-8B~\cite{arxiv2025_qwen3}, to condense the detailed descriptions.
The weights of $\mathcal{L}_{O}$ and $\mathcal{L}_{3D}$ in Eq.~\ref{total_loss} are $\lambda_1=0.01$ and $\lambda_2=10$, respectively.
For sampling, we utilize classifier-free guidance (CFG) with a scale of 2 for our motion condition and the official scale of 4 for the text input.
All training and evaluation are conducted on a single NVIDIA A100 GPU.
The total training process takes approximately 20 hours, while the average automated inference latency is about 30s/sample, calculated by ignoring the case of manual intervention (point-tracking workflow) and averaging the costs of VLM-based tool selection (4.08s), joint estimation (25.77s), and conditional motion generation (0.51s).
More details are provided in the Appendix.

\paragraph{Baselines}
As few existing methods directly target our task, we consider three categories of representative baselines that can be adapted to character-video-driven human reenactment. 
The first category, termed VLM+T2M, leverages large vision-language models, Qwen3.5-27B~\cite{arxiv2026_qwen3.5}, to generate detailed descriptive prompts for motion reenactment, which are then fed into text-to-motion models, including:
(1) MoMask++~\cite{NIPS2025_snapmogen}: the base motion generation model in our framework with approximately 31M parameters;
(2) HY-Motion~\cite{arxiv2025_hymotion}: a large-scale industrial model with 1.4B parameters trained on massive data (over 3000 hours).
The second category, termed Video2SMPL(X), leverages powerful vision encoders or large video diffusion models to directly regress the human template parameters~(SMPL(X)) for reconstructing 3D human motion based on character video, including GENMO~\cite{ICCV2025_GENMO} and EchoMotion~\cite{ICLR2026_echomotion}.
The third type, termed VT+Mocap, first utilizes a video-diffusion-model-based method, DisMo~\cite{NIPS2025_DisMo}, to performs motion transfer in the video domain, and then uses a state-of-the-art video-to-human motion model, GVHMR~\cite{SA2024_world}, to reconstruct the reenactment 3D human motion form video.
Since the released DisMo is based on an image-to-video model (CogVideoX-5B-I2V~\cite{ICLR2025_cogvideox}) that requires a first frame containing plausible human pose for further motion transfer, we use Qwen-Image~\cite{arxiv2025_qwen_image} to generate this initial condition.

From the perspective of conditioning inputs, AnyAct takes a reference video and a brief motion description as input.
The VFE independently processes the reference video to extract the sparse local 2D joint trajectory.
The trajectory, together with the motion description, are then fed into the conditional motion generation stage to produce the reenacted 3D human motion.
For the VLM+T2M baselines, the system takes only the reference video as input.
The VLM first generates a detailed textual description of the motion to be reenacted, which is subsequently used as the input to the text-to-motion model to generate the corresponding 3D human motion.
For the Video2SMPL(X) baselines, GENMO takes only the reference video as input, whereas EchoMotion additionally uses a brief motion description.
Both methods directly output the reenacted 3D human motion.
For the VT+Mocap baseline, the inputs consist of a reference video, an image-editing instruction, and a motion description.
The first frame of the reference video and the editing instruction are first fed into Qwen-Image to generate a character image with a plausible human pose.
This image is then used as the first-frame condition for DisMo, together with the reference video and motion description, to generate a human reenactment video.
Finally, GVHMR reconstructs the reenacted 3D human motion from the generated video.

\begin{figure*}[!tb]
    \centering
    \includegraphics[width=1.0\linewidth]{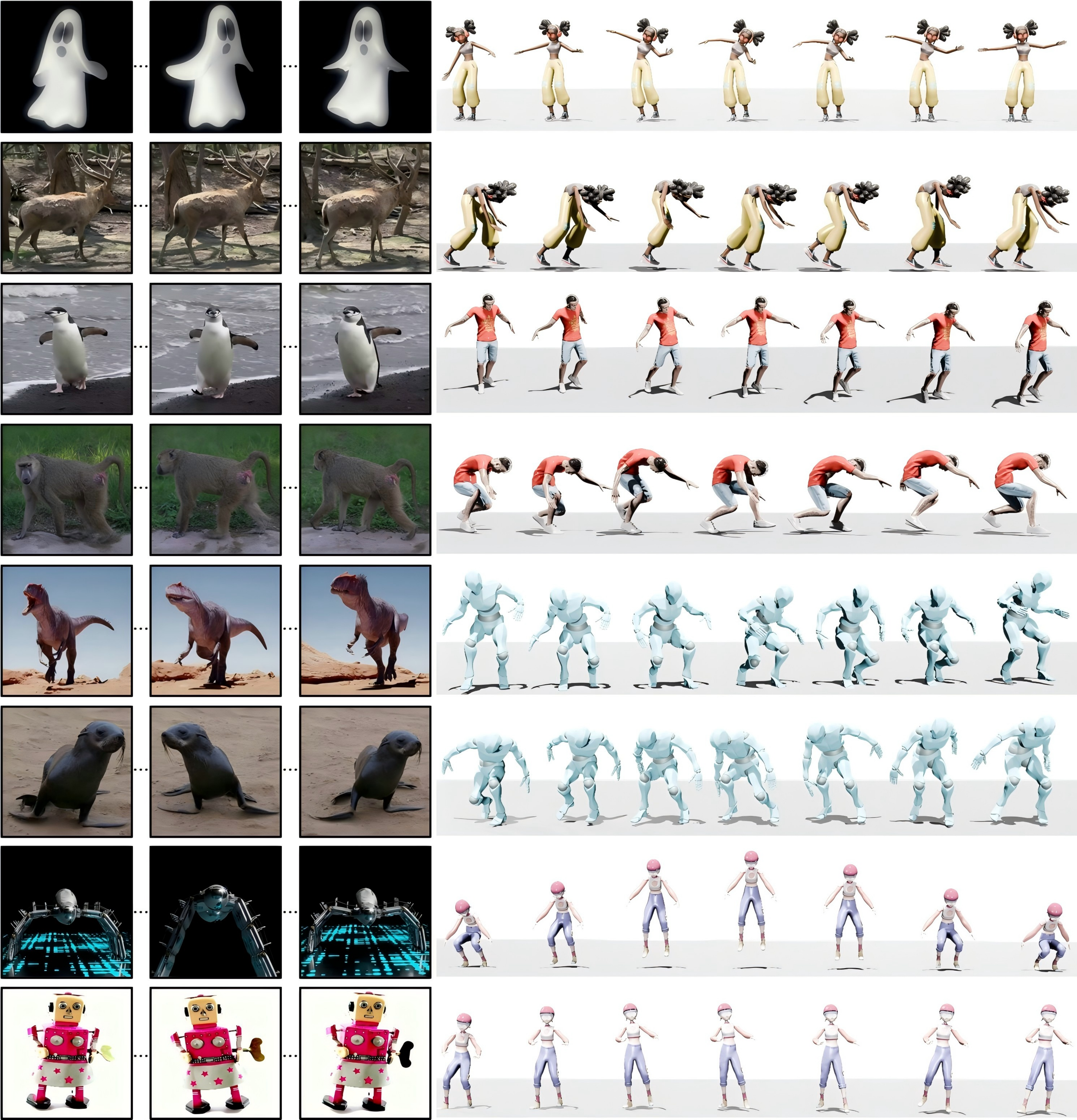}
    \caption{\textbf{Result Gallery.}
    (1) dancing with side-to-side swaying, following the rhythm of the ghost,
    (2) deer-like walking,
    (3) penguin-like walking,
    (4) monkey-like walking,
    (5) dinosaur-like walking,
    (6) seal-like walking (with side-to-side swaying),
    (7) mechanical-spider-like in-place jumping,
    (8) toy-robot-like walking (with side-to-side swaying).}
    \label{fig:more_result}
\end{figure*}

\begin{figure*}[!tb]
    \centering

    \begin{tikzpicture}
        \node[anchor=south west, inner sep=0] (img) at (0,0)
        {\includegraphics[width=1\linewidth]{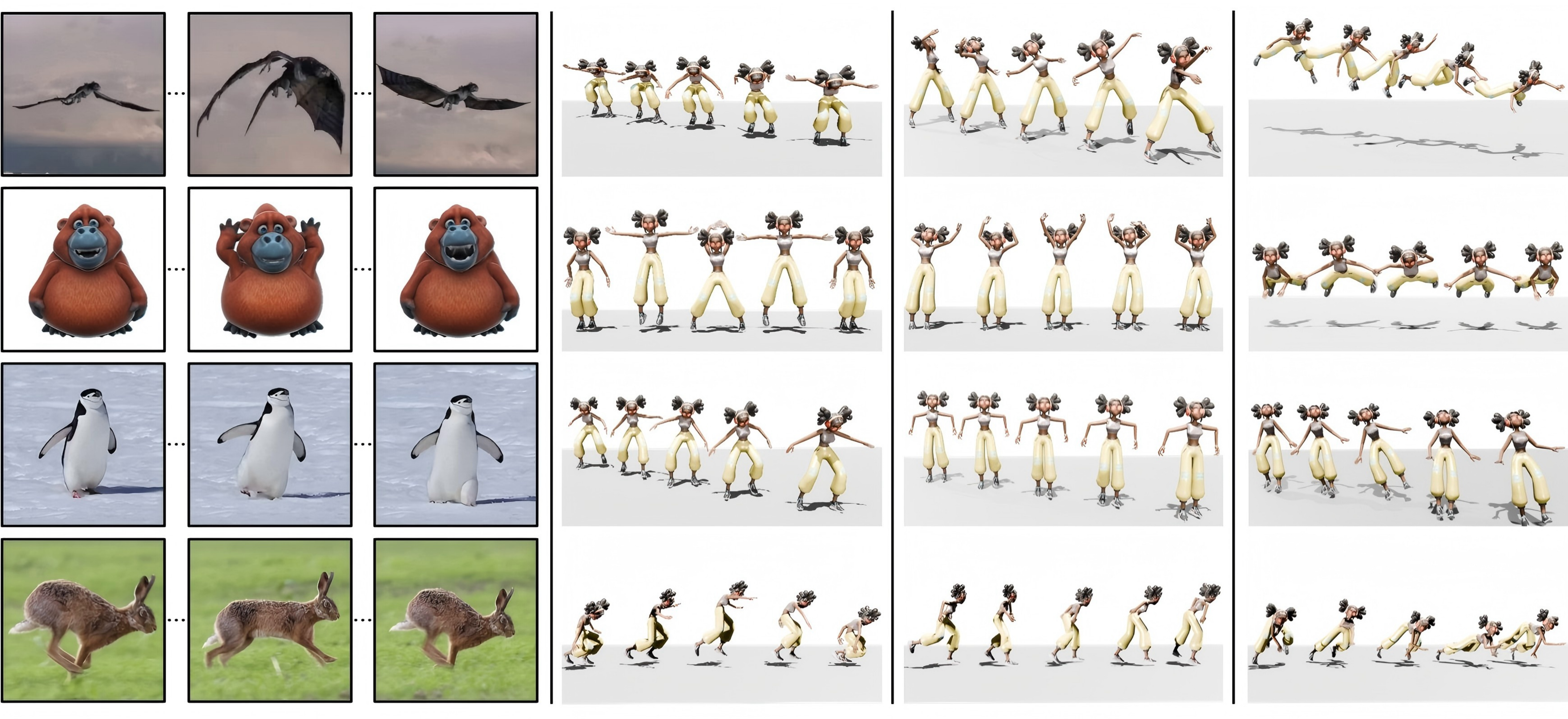}};

        \begin{scope}[x={(img.south east)}, y={(img.north west)}]
        \node[anchor=north, font=\bfseries\small] at (0.0626,1.03) {Reference Video};
            \node[anchor=north, font=\bfseries\small] at (0.405,1.03) {AnyAct (Ours)};
            \node[anchor=north, font=\bfseries\small] at (0.636,1.03) {VLM + HY-Motion};
            \node[anchor=north, font=\bfseries\small] at (0.833,1.03) {EchoMotion};
        \end{scope}
    \end{tikzpicture}
    \vspace{-6.5mm}
    \caption{\textbf{Qualitative comparison of AnyAct against VLM+HY-Motion and EchoMotion.} Based on the reference videos, human should perform: (1) monster-like flying, (2) cartoon bear-like jumping, (3) penguin-like walking, and (4) rabbit-like bounding.
    The results demonstrate that our method achieves superior reenactment quality compared to the other two baselines, while preserving the plausibility of the motion.}
    \label{fig:vis_comparison}
\end{figure*}
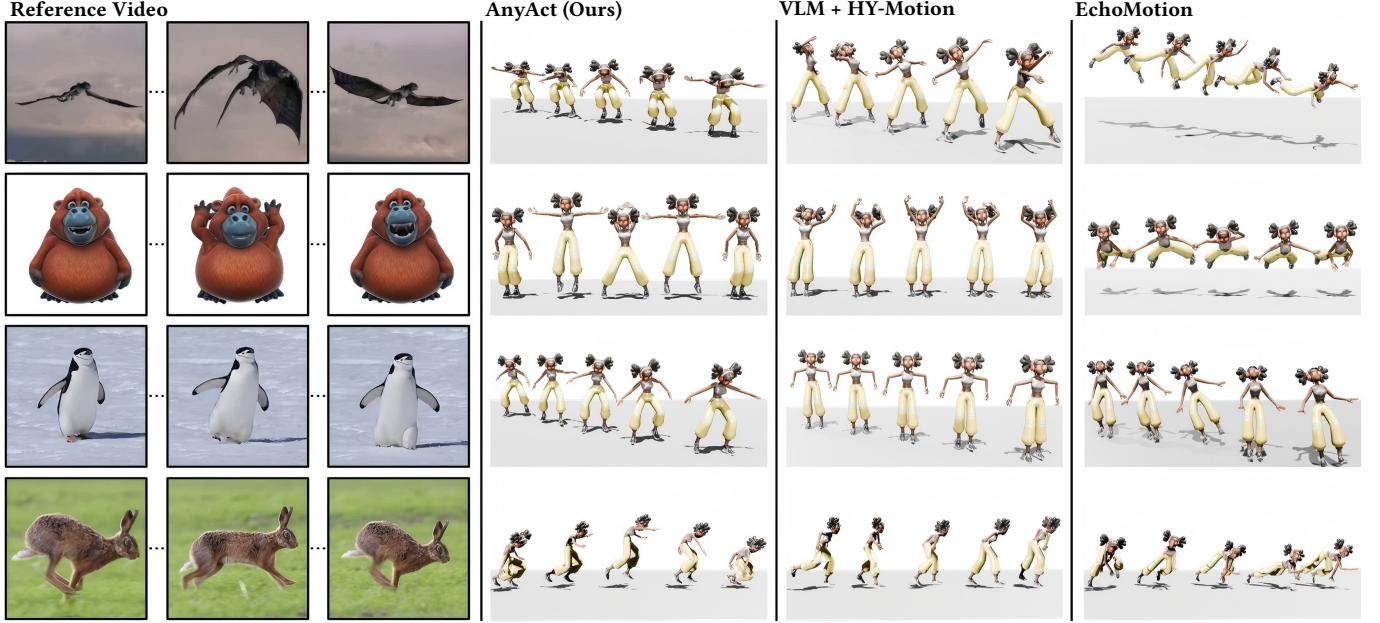

\begin{figure}[!tb]
    \centering
    \includegraphics[width=1.0\linewidth]{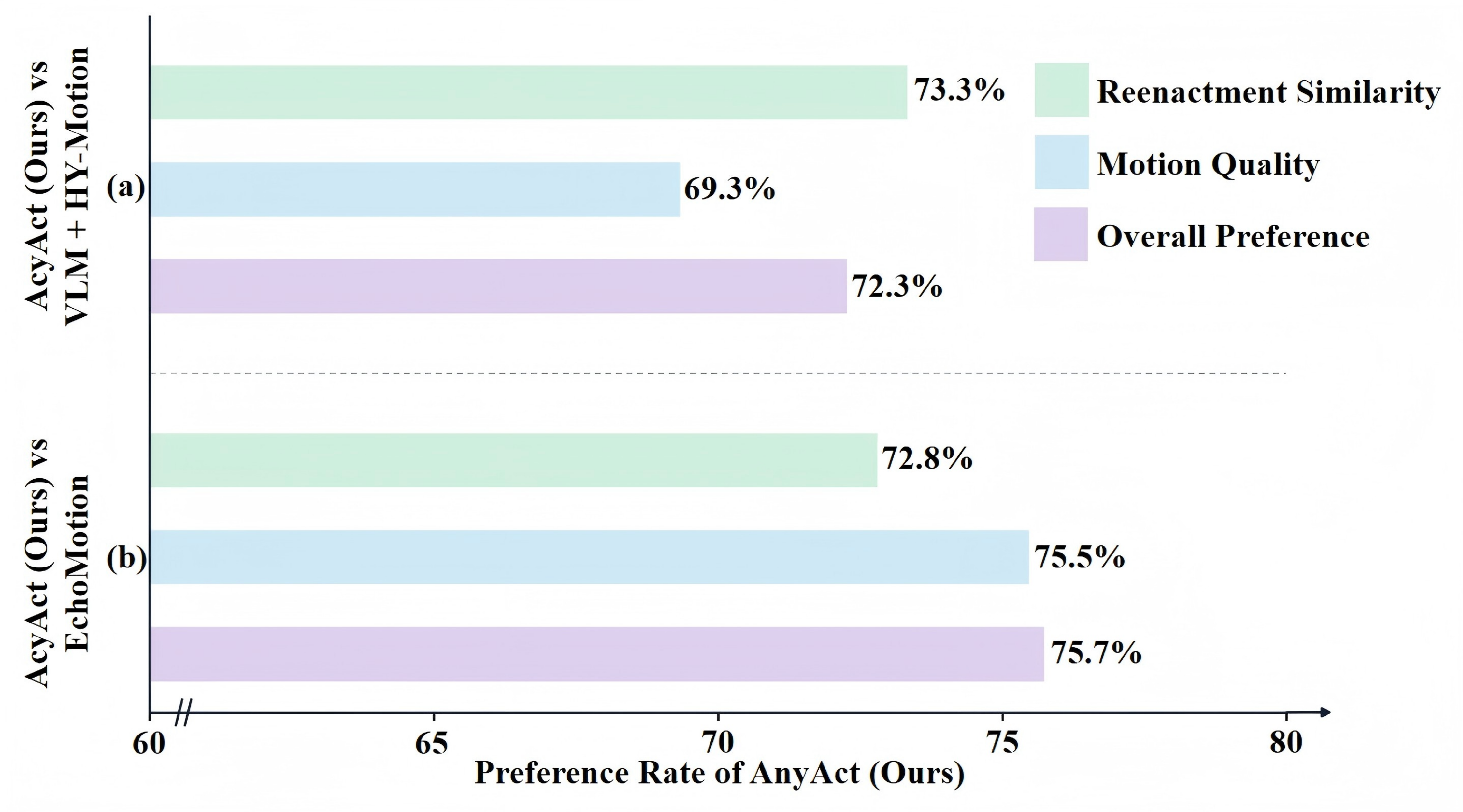}
    \vspace{-6mm}
    \caption{\textbf{Result of the user study.}
    We report the preference rates of our \name in pairwise comparisons against (a) VLM+HY-Motion and (b) EchoMotion.
    Participants evaluated the generated motions based on \textit{Reenactment Similarity}, \textit{Motion Quality}, and \textit{Overall Preference}, respectively.
    Our method consistently outperforms both baselines across all criteria.}
    \vspace{-6mm}
    \label{fig:user_study}
\end{figure}

\paragraph{Evaluation Metrics}
We first evaluate \name and all baselines using a state-of-the-art VLM (Google/Gemini-3) following recent works~\cite{arxiv2025_motionpersona, ICLR2026_quest, arxiv2026_t2mbench}. The evaluation spans four dimensions: \textit{Motion Semantic Consistency (MSC)}, \textit{Motion Rhythm Consistency (MRC)}, \textit{Posture Adaptation (PA)}, and \textit{Human Motion Plausibility (HMP)}. The first three measure reenactment consistency with the reference video, while the last assesses the kinematic plausibility of the generated motion. Following~\cite{ICCV2025_MotionMillion, arxiv2026_t2mbench}, the VLM scores each dimension on a 0-10 scale. To ensure a fair comparison across methods with different motion representations, we convert all outputs to SMPL meshes and render them as videos with a ground plane.
To objectively evaluate articulation fidelity, we measure the similarity of corresponding sparse local 2D joint motions between the input condition and the generated 3D human motion.
Specifically, for each generated 3D motion, we sample 100 fixed camera views, re-project 3D joints to each 2D camera plane, apply the same root-centering/scale-normalization as the input condition, and compute the average L2 distance between the condition and re-projected joints for each view.
To account for monocular camera ambiguity, we report the minimum value (JD-Min) among these views.
More details are provided in the Appendix.
We further report physical quality metrics, including Jitter, Foot Skating Ratio (FSR), Foot Floating (FFL), and Foot Sliding Distance (FSD)~\cite{SA2025_stablemotion}. Since these metrics are sensitive to scale, frame rate, and contact definitions, direct comparison across methods trained under different settings is unreliable. We therefore follow the official MoMask++ implementation and evaluate them only on methods built on the MoMask++ architecture. Instead of reporting absolute values, we report relative results with respect to the adapter-free baseline (VLM+MoMask++), denoted as \textit{R-Jitter}, \textit{R-FSR}, \textit{R-FFL}, and \textit{R-FSD}. We find these relative metrics to be better aligned with VLM-based evaluation and human perception.

\begin{table}[!t]
    \centering
    \small
    \setlength{\tabcolsep}{5pt}
    \renewcommand{\arraystretch}{1.0}
    \caption{\textbf{Quantitative comparison of AnyAct with baselines adapted to our setting.}The best and second-best results are shown in \textbf{bold} and \underline{underline}, respectively.}
    \label{tab:quantitative_comparison}
    
    \begin{tabular}{lccccc}
        \toprule
        \textbf{Method} & \textbf{MSC}$\uparrow$ & \textbf{MRC}$\uparrow$ & \textbf{PA}$\uparrow$ & \textbf{HMP}$\uparrow$ & \textbf{JD-Min}$\downarrow$\\
        \midrule
        VLM + MoMask++   & 5.544 & 5.663 & 5.525 & 9.033 &0.345 \\
        VLM + HY-Motion  & 5.794 & 6.025 & 5.881 & \textbf{9.181} &0.372 \\
        \midrule
        GENMO            & 5.563 & 5.775 & 5.644 & 7.644 &0.320 \\
        EchoMotion       & \underline{6.213} & \underline{6.313} & \underline{6.294} & 8.325 &\underline{0.296} \\
        \midrule
        VT+Mocap & 6.036 & 6.144 & 6.078 & 8.650 &0.341 \\
        \midrule
        \textbf{AnyAct (Ours)} & \textbf{6.619} & \textbf{6.763} & \textbf{6.700} & \underline{9.056} &\textbf{0.251}\\
        \bottomrule
    \end{tabular}
    \vspace{-3mm}
\end{table}

\begin{table*}[t]
  \centering
  \caption{\textbf{Ablation study of core designs in AnyAct.}
  We evaluate various core design choices of our \name across multiple VLM-based and physical metrics.
  The best and second-best scores for each metric are \textbf{bolded} and \underline{underlined}, respectively.}
  \label{tab:ablation}
  \renewcommand{\arraystretch}{0.95} 
  \scalebox{0.99}{
  \begin{tabular}{ll cccccccc}
    \toprule
    && \textbf{MSC} $\uparrow$ & \textbf{MRC} $\uparrow$ & \textbf{PA} $\uparrow$ & \textbf{HMP} $\uparrow$ & \textbf{R-Jitter} $\downarrow$ & \textbf{R-FSR} $\downarrow$ & \textbf{R-FFL} $\downarrow$ & \textbf{R-FSD} $\downarrow$ \\
    \midrule
    \multicolumn{2}{c}{\textbf{AnyAct-Base (Ours)}} & \textbf{6.619} & \textbf{6.763} & \textbf{6.700} & \textbf{9.056} &\textbf{0.058} &\underline{0.013} &\textbf{0.007} &\textbf{0.001} \\
    \midrule
    \multirow{1}{*}{(a) Feature Extraction} & ViTPose++ Only & 6.125 & 6.306 & 6.344 & 8.650 & 0.200 & 0.032 & 0.044 & 0.014 \\
    \midrule
    \multirow{2}{*}{(b) 3D-to-2D Training} & 3D Joint Input & 5.719 & 6.006 & 5.581 & \underline{9.044} & \underline{0.079} & \textbf{0.002} & 0.021 & 0.003 \\
    & w/o $\mathcal{L}_{3D}$ & 6.344 & 6.513 & \underline{6.444} & 8.681 & 0.185 & 0.033 & 0.044 & 0.008 \\
    \midrule
    \multirow{3}{*}{(c) Local Motion Focused} & w/o 3D-GA & 6.444 & 6.581 & 6.488 & 8.900 & 0.142 & 0.023 & 0.045 & 0.006 \\
    & w/o $\mathcal{L}_{O}$ & \underline{6.488} & \underline{6.688} & \underline{6.444} & 9.031 & 0.082 & 0.016 & \underline{0.010} & 0.004 \\
    & Freeze the 3D-GA & 6.375 & 6.650 & 6.406 & 9.019 & 0.102 & 0.016 & 0.013 & \underline{0.002} \\
    \bottomrule
  \end{tabular}
  }
\end{table*}

\paragraph{Quantitative Comparison}
Tab.~\ref{tab:quantitative_comparison} evaluates \name against several state-of-the-art baselines adapted to our task.
For VLM-based evaluation, \name achieves the best performance in MSC, MRC, and PA, demonstrating its superior ability to capture the motion characteristics of source characters.
While our HMP is slightly lower than that of VLM+HY-Motion, this is primarily due to the inherent disparity in base model capacity.
HY-Motion is an industrial-scale model, exceeding MoMask++ by over 45$\times$ in parameters and 65$\times$ in data volume.
Consequently, HY-Motion naturally outperforms MoMask++ across all metrics when using the same VLM.
Notably, despite being built upon the much smaller base model, \name successfully bridges this performance gap through its meticulous design, vastly surpassing VLM+HY-Motion in reenactment precision.
Furthermore, while EchoMotion leverages the extensive world knowledge of the video diffusion model Wan2.2-5B \cite{arxiv2025_wan} to achieve a degree of cross-species generalization and yields the second-best performance in reenactment consistency, it suffers from lower motion quality (HMP) compared to methods trained on high-fidelity motion data.
Besides, the VT+Mocap method, based on open-source models, results in poor performance across all metrics, indicating that video-domain transfer plus human reconstruction does not reliably solve our non-human reenactment setting.
This will be further discussed in the Appendix, where we replace DisMo with a more advanced closed-source video generation and motion transfer model, Seedance 2.0.
For the reference-articulation fidelity, our AnyAct also achieves the lowest JD, reflecting its superior condition alignment.

\paragraph{Qualitative Comparison}
Fig.~\ref{fig:vis_comparison} displays the qualitative comparison of our \name against VLM+HY-Motion and EchoMotion (the top-performing methods in their respective technical categories).
The results show that VLM+HY-Motion can produce plausible human motions while failing to align them with the dynamics of characters in reference videos.
For motions generated by EchoMotion, although their dynamics partially follow the source characters in the videos, they exhibit poor kinematic quality (e.g., several floating artifacts and unnatural poses).
In contrast, our \name preserves the kinematic plausibility of the generated motions while capturing the essential dynamics of characters in the reference videos.
More visual reenactment results of our method are provided in Fig.~\ref{fig:more_result}.

\paragraph{User Evaluation}
To further assess perceptual quality, we conduct a user study comparing \name against VLM+HY-Motion and EchoMotion separately.
Specifically, we select 30 reference character videos from our benchmark and run each method to output the corresponding human reenactment motions.
For visualization of the generated motions, we follow the same rendering pipeline as the VLM-based evaluation.
Subsequently, given a reference video, we ask users to perform pairwise comparisons (\name vs VLM+HY-Motion/EchoMotion), selecting the superior output in terms of \textit{Reenactment Similarity}, \textit{Motion Quality}, and \textit{Overall Preference}, respectively.
A total of 50 participants took part in our user study, contributing 4500 valid votes.
The user study results in Fig.~\ref{fig:user_study} reveal that \name consistently outperforms the two state-of-the-art baselines (VLM+HY-Motion and EchoMotion), securing an approval rating exceeding 70\% across both Reenactment Similarity and Overall Preference, underlining the perceptual superiority of our framework.


\begin{figure*}[!tb]
    \centering
    \includegraphics[width=1\linewidth]{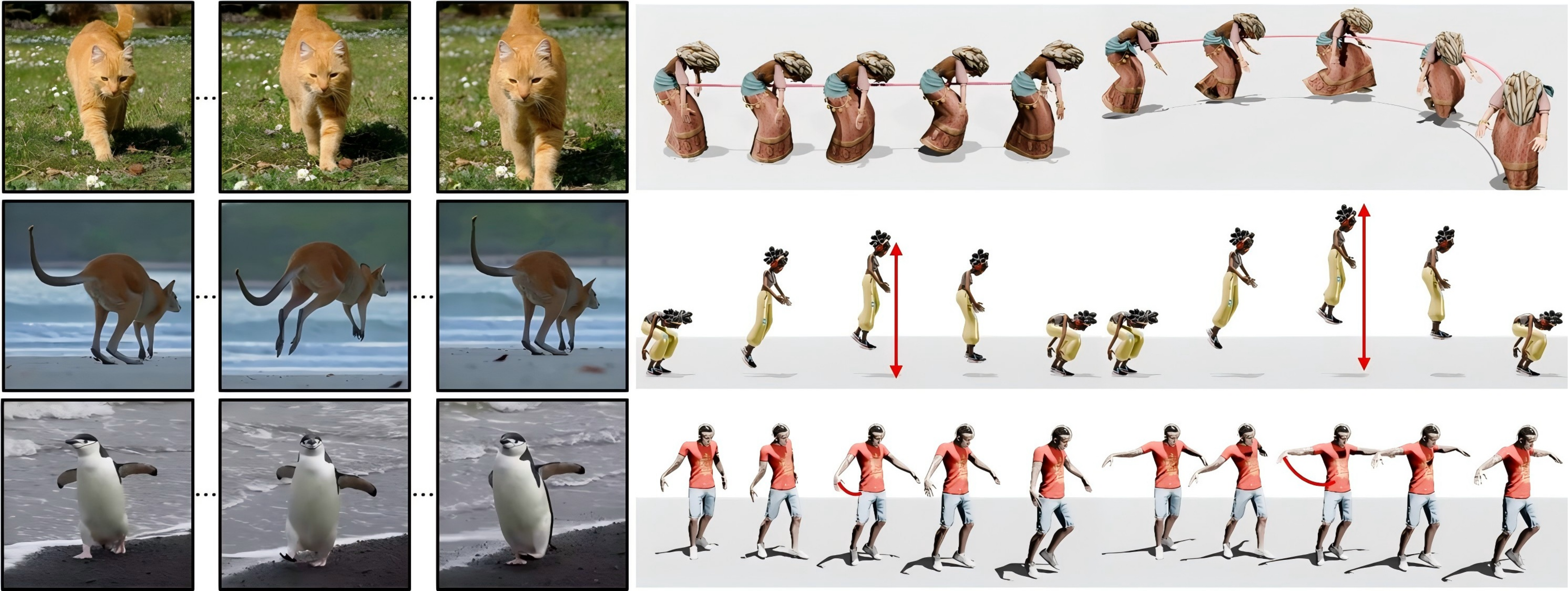}
    \caption{\textbf{Trajectory control and intuitive editing.}
    (1) cat-like walking with forward and half-circle trajectory control,
    (2) editing the height of human when perform kangaroo-like jumping,
    (3) editing the arm spread of human when perform penguin-like walking.
    }
    \vspace{-1mm}
    \label{fig:traj_control&edition}
\end{figure*}

\begin{figure*}[!tb]
    \centering
    \includegraphics[width=1\linewidth]{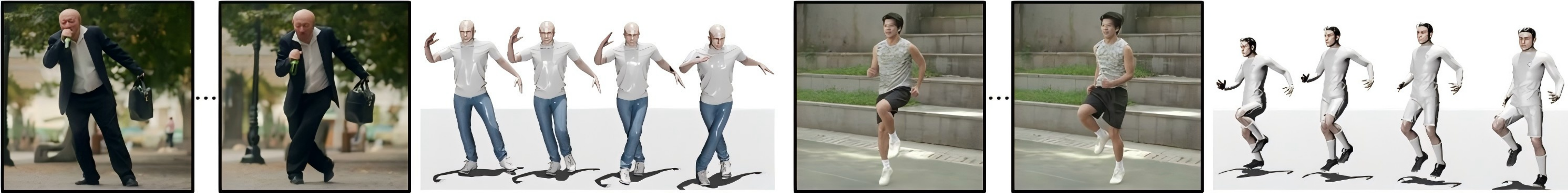}
    \caption{\textbf{Generalization to the reenactment of human characters.}
    Although our \name does not aim to achieve the same level of absolute reconstruction as human-centric mocap-based methods, it still demonstrates the ability to perform reliable reenactment of human characters from monocular videos.
    }
    \vspace{-1mm}
    \label{fig:generalize2human}
\end{figure*}

\begin{figure*}[!tb]
    \centering
    \includegraphics[width=1\linewidth]{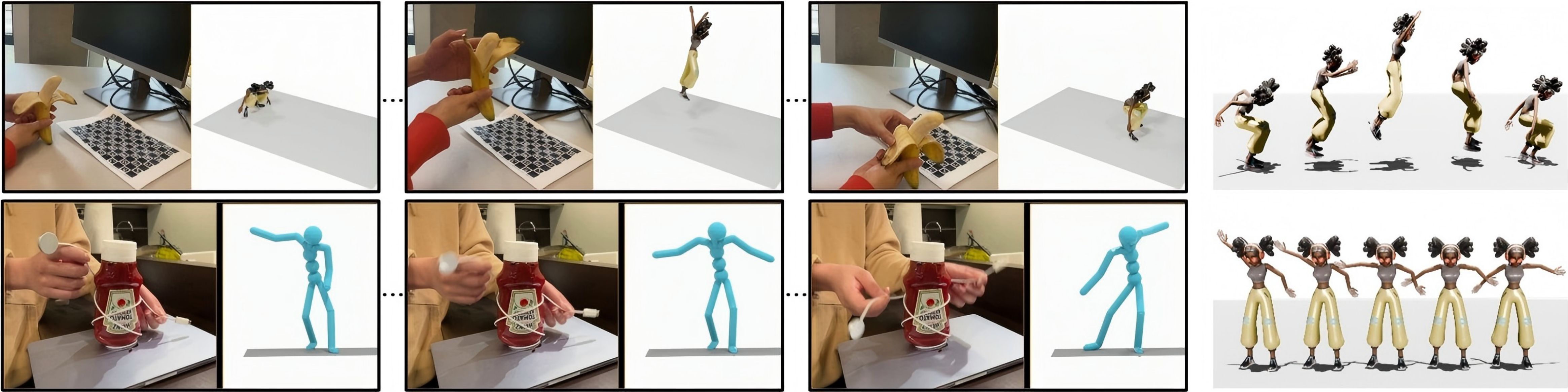}
    
    \caption{\textbf{Results of DancingBox.}
    Reference videos are obtained from the DancingBox project page, where the source captures and results are concatenated together.
    Our results (in the right part) confirms that \name can effectively recovers the essential dynamics of physical proxies as DancingBox.
    }
    \vspace{-1mm}
    \label{fig:compare_dancingbox}
\end{figure*}

\begin{figure*}[!tb]
    \centering
    \includegraphics[width=1\linewidth]{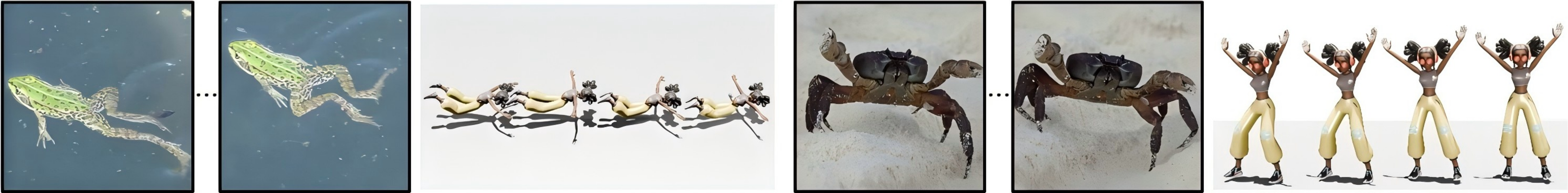}
    \caption{\textbf{Limitation.}
    Left: \name struggles to generate plausible reenactments for motions far outside the training distribution, such as frog-like swimming involving rapid leg kicks and a rare prone posture.
    Right: Actions like crab-like sideways walking with rapid multi-leg movements and pixel-level similarities with neighboring points can cause CoTracker3 tracking failures, resulting in noisy 2D features that impair subsequent motion generation.
    }
    \label{fig:Limitation}
\end{figure*}

\subsection{Ablation Study}

\paragraph{Motion Feature Extraction}
We ablate the proposed Versatile Feature Extractor (VFE) by using only ViTPose++ to predict sparse joints for all inputs. As shown in Tab.~\ref{tab:ablation}(a), this simplification leads to clear degradation in both VLM-based and physical metrics. The result confirms that a single estimator is insufficient for the diverse character topologies considered in our task and validates the need for the model-ensemble-based extraction strategy in VFE.

\paragraph{Progressive 3D-to-2D Training}
We first examine a direct 3D-joint-based control variant of \name. Specifically, we replace the 2D joint estimators with 3D predictors~\cite{CVPR2026_FMPose3D}, and follow DancingBox~\cite{CHI2026_DancingBox} by using a 4D reconstruction model~\cite{ICLR2026_Pi3} to lift the predicted 2D points into 3D. We then train a 3D local adapter under the same setting. As shown in Tab.~\ref{tab:ablation}(b), this variant performs substantially worse than our 2D-cue-based formulation, mainly because 3D joint estimation from monocular character videos is significantly less reliable than 2D motion cue extraction.
We also ablate the progressive training strategy by removing the 3D-to-2D alignment loss. This causes clear performance drops in both condition following and motion quality, indicating that direct training on projected 2D cues alone provides insufficient structural guidance. These results support our progressive 3D-to-2D training scheme, which transfers depth-aware motion knowledge from the stronger 3D branch to the 2D branch.

\paragraph{Local Motion Focused Learning}
Tab.~\ref{tab:ablation}(c) validates the importance of decoupling local motion control from global trajectory modeling. Removing the auxiliary global trajectory branch degrades performance, while adding the orthogonality regularizer further improves both condition following and motion quality. By contrast, freezing the global trajectory branch during 2D training causes a noticeable drop, suggesting that effective conditioning requires joint adaptation between the global branch and the 2D local branch.


\subsection{Application and Discussions}
\label{subsec:Application and Discussion}
\paragraph{Trajectory Control and Intuitive Editing}
Beyond generating an initial human reenactment, \name also supports downstream control and editing. Following~\cite{ICCV2025_maskcontrol}, the generated motion can be controlled by user-specified trajectories. In addition, animators can apply intuitive edits to better match desired performances, such as adjusting the jump height of kangaroo-like motion by scaling the vertical root offset or controlling the arm spread of penguin-like walking by modifying the angle between the arm and the forward axis. Representative results are shown in Fig.~\ref{fig:traj_control&edition}.

\paragraph{Generalization to Human Videos}
Although \name is designed for non-human character videos, it also generalizes to human videos. As shown in Fig.~\ref{fig:generalize2human}, it still produces reliable reenactments in this setting, suggesting that the proposed motion representation and conditioning strategy can extend to the human domain.

\paragraph{Discussion of DancingBox}
As shown in Fig.~\ref{fig:compare_dancingbox}, \name can recover the essential dynamics of the physical-proxy examples presented on the DancingBox website. However, the two methods target different settings: DancingBox relies on constrained capture setups with explicit ground cues, user interaction, and per-sample hyperparameter adjustment, whereas our method is designed for monocular character videos captured in unconstrained environments.

\paragraph{Limitation}
\name is limited by the motion coverage of its base generator~\cite{NIPS2025_snapmogen}, which is trained on a relatively small mocap dataset. As a result, it may struggle with motions far outside the training distribution, such as frog-like swimming with rapid leg kicks and rare prone-body configurations (Fig.~\ref{fig:Limitation}).
Furthermore, as shown in our video demo, our method sometimes produces results with artifacts such as jitter and foot sliding.
However, these may be alleviated by using a stronger base generator, such as recent large-scale motion models~\cite{arxiv2025_hymotion, arxiv2026_kimodo}.
Another limitation lies in motion cue extraction. In out-of-domain cases that rely on point tracking, CoTracker3 may fail under rapid motion or strong local visual similarity. Crab-like sideways walking is one such example, where fast and repetitive multi-leg motion can corrupt the extracted sparse 2D cues and degrade reenactment quality.

\section{Conclusion}
We studied the problem of deriving an initial human reenactment directly from a monocular video of a non-human character. To address this setting, we introduced \name, a framework that formulates character-video-driven human reenactment as conditional human motion generation from transferable sparse local 2D motion cues. Our formulation is enabled by three key designs: human-motion-only supervision via augmented 3D-to-2D projection, progressive 3D-to-2D training for ambiguity reduction, and global-local motion decoupling for reliable local motion control. 
Experiments on our benchmark demonstrate the effectiveness of our \name, while ablation studies validate the contribution of each core design.


\bibliographystyle{ACM-Reference-Format}
\bibliography{main}


\end{document}